\documentclass[11pt]{article}

\usepackage{acl}
\usepackage{times}
\usepackage{latexsym}
\usepackage[T1]{fontenc}
\usepackage[utf8]{inputenc}
\usepackage{microtype}
\usepackage{inconsolata}
\usepackage{graphicx}
\usepackage{booktabs}
\usepackage{amsmath}
\usepackage{amssymb}
\usepackage{multirow}
\usepackage{url}
\usepackage{booktabs}
\usepackage{multirow}
\usepackage{graphicx}

\newcommand{\method}{\textsc{HiEvi-RAG}}
\newcommand{\eviagent}{\textsc{EviAgent}}

\title{\method: Hierarchical Evidence-Driven Reasoning for Long Document Understanding}

\author{
Junyu Xiong$^{1}$
\quad
Yonghui Wang$^{1}$
\quad
Rongjian Gu$^{1}$ 
\quad
Chenyu Liu$^{2}$
\\
\textbf{Bing Yin}$^{2}$
\quad
\textbf{Wengang Zhou}$^{1}$\thanks{Corresponding author.}
\quad
\textbf{Houqiang Li}$^{1}$ \\
$^{1}$University of Science and Technology of China \\
$^{2}$iFLYTEK Research \\
\texttt{
xiongjyu@mail.ustc.edu.cn,
zhwg@ustc.edu.cn
}
}

\begin{document}
\maketitle

\begin{abstract} 
Retrieval-Augmented Generation (RAG) streamlines long-document understanding by leveraging retrieval mechanisms to restrict input images to a highly curated subset. 
However, existing multimodal RAG pipelines primarily face two critical challenges: 
first, standard semantic similarity retrievers frequently fetch topically overlapping yet answer-void distractor pages that mislead downstream generation; 
second, rigid single-pass pipelines heavily depend on initial retrieval success, where any omission of core evidence inevitably causes cascading errors. 
To address these challenges, we introduce \textbf{\method}, a hierarchical, evidence-driven multimodal RAG framework for closed-domain document understanding. 
\method\ systematically factorizes complex queries into a cooperative four-stage pipeline: 
(1) hierarchical question decomposition to break multi-hop root queries into atomic child questions; 
(2) coarse visual page retrieval leveraging a multimodal retriever to fetch candidate pages based on semantic similarity;  
(3) fine-grained page verification via \eviagent, a specialized multi-page verifier trained with GRPO to execute cross-page reasoning over multi-image blocks; and 
(4) memory-guided iterative generation that leverages accumulated sub-question context to execute multi-round, dynamic reasoning over the prioritized sequence.
Extensive evaluations across four benchmarks demonstrate the robust efficacy and synergy of our framework, which significantly outperforms existing open-source baselines and exceeds the strongest reported baseline by an average of 8.05\% in accuracy.
\end{abstract}

\section{Introduction}
Closed-domain document understanding has become an indispensable core capability for practical AI systems deployed across various sectors, including science, finance, enterprise knowledge management, and technical support~\cite{hui2024uda,suri-etal-2025-visdom,deng-etal-2025-longdocurl,dong2024encoding,xia2024vision}. In these scenarios, users typically pose queries over a proprietary document collection, where the answers are sparsely distributed across hundreds of pages containing diverse and rich information such as plain text, tables, charts, and illustrations~\cite{wang2025mmlongbench,chia-etal-2025-longdoc}. However, due to the prohibitive number of images in these closed-domain documents, feeding the entire collection into a model in an end-to-end manner easily exceeds the context window limits of existing LLMs. 

To address this bottleneck, a growing body of work on Retrieval-Augmented Generation (RAG) has emerged, which leverages retrieval mechanisms to restrict the number of input images to a highly curated subset prior to answer generation~\cite{lewis2020retrieval,cho2024m3docrag,suri-etal-2025-visdom,xiong2026priorzero}. Initially, to bypass traditional OCR limitations and capture multimodal information, ColPali~\cite{faysse2025colpali} extends late interaction to rendered page images, while M3DocRAG~\cite{cho2024m3docrag} establishes this method as the core interface for multi-page QA. Subsequently, research shifted toward filtering distractors and optimizing ranking precision: MDocAgent~\cite{han2025mdocagent} and SimpleDoc~\cite{jain-etal-2025-simpledoc} refine results via module collaboration or iterative cascading; RagVL~\cite{chen-etal-2025-vlm} and MM-R5~\cite{xu2025mmr5} employ instruction-tuning or reinforcement learning to train strong VLM rerankers; and MoLoRAG~\cite{wu-etal-2025-molorag} leverages page topology graphs for logic-aware traversal. Concurrently, DocAgent~\cite{sun-etal-2025-docagent} and ALDEN~\cite{yang2026alden} utilize memory feedback or active exploration protocols to gather evidence dynamically, while DMAP~\cite{fu2026dmap} constructs human-aligned structural maps for global document comprehension. Despite these advancements, two challenges remain: first, relying on semantic similarity misleads models with topically related yet answer-void pages; second, single-pass pipelines rigidly depend on initial retrieval, where any omission of core evidence makes correct generation impossible.

To address these challenges, we propose \textbf{\method}, a four-step framework for long-document understanding. First, \textbf{hierarchical question decomposition} breaks down the complex, multi-hop root question into multiple atomic sub-questions, reducing retrieval difficulty while assisting the root question's resolution from different logical levels. Second, \textbf{coarse-grained page retrieval} utilizes a multimodal retriever (e.g., ops-ColQwen3) to fetch the top-$k$ pages based on semantic similarity for each question, including both root and sub-questions. Third, for \textbf{fine-grained evidence verification}, inspired by DocR1~\cite{xiong2025docr1}, we employ GRPO~\cite{shao2024deepseekmath} to train \textbf{\eviagent}, an evidence-aware multi-page reasoning model capable of cross-page reasoning and grounding over multi-image inputs to pinpoint question-relevant evidence. In this stage, for sub-questions, we strictly retain only the evidence pages identified by \eviagent; for the root question, \eviagent's verified evidence pages are prioritized at the front, while non-evidence pages follow in order of their initial semantic similarity, thereby achieving precise reranking. Fourth, during \textbf{memory-guided iterative generation}, an empty memory is initialized. The sub-questions and their corresponding evidence pages are fed into the model, with the sub-questions, generated reasoning traces, and intermediate answers subsequently accumulated in the memory. For the root question, the model takes the root query, the accumulated memory, and a batch of images as input. If the current memory and image batch are insufficient to resolve the question, the system appends the latest reasoning trace to the memory and fetches the next batch of images for iterative reasoning.

To verify the effectiveness of our approach, we conduct extensive experiments and analyses across four distinct datasets. The results demonstrate that our method significantly outperforms existing open-source baselines, exceeding the previous state-of-the-art by an average of 8.05\%. Furthermore, we perform comprehensive ablation studies to validate the individual efficacy of each proposed module.

Our contributions can be summarized as follows:
\begin{itemize}
\item We propose \method, a four-step multimodal RAG framework for long-document QA that unifies hierarchical question decomposition, coarse-to-fine retrieval, and memory-guided iterative generation.
\item We introduce a fine-grained verification mechanism via \eviagent, which performs cross-page reasoning over multi-image inputs to precisely ground answer-supporting evidence.
\item We formulate a memory-guided iterative generation process that utilizes sub-question reasoning traces to assist the root query while dynamically updating an explicit memory across multiple rounds to eliminate cascading retrieval errors.
\end{itemize}

\section{Related Work}
\subsection{Document Visual Question Answering}
Advancements in DocVQA have bifurcated into OCR-based and OCR-free paradigms. 
OCR-based methods convert pages into structured textual and spatial tokens to feed downstream reasoning frameworks~\cite{tang2023udop, wang2024docllm, zhu2025laytokenllm}, yet their performance remains inherently bounded by compounding parsing errors and the serialization loss of visual context. 
Conversely, OCR-free approaches directly train VLMs on raw document images, evolving from architectural scaling for resolution management~\cite{ye2023ureader, feng2023docpedia, hu2024docowl2} to evidence-aware reasoning via reinforcement learning~\cite{xiong2025docr1, zheng2026docvstar}. 
While these methods substantially enhance single- or multi-page reader execution, effectively navigating and distilling massive long-document repositories remains a largely orthogonal challenge. 

\subsection{Retrieval-Augmented Generation}
Multimodal RAG methodologies have transitioned from coarse-grained text-retrieval extensions to visual-centric, 
layout-aware document discovery. Early pipelines leverage visual retrievers and VLM-based rerankers to screen candidate pages independently based on superficial similarity scoring~\cite{faysse2025colpali, cho2024m3docrag, chen-etal-2025-vlm, xu2025mmr5}. 
More recent efforts incorporate iterative multi-agent navigation, memory feedback, or topological page graphs to capture structural inter-page logic~\cite{wu-etal-2025-molorag, sun-etal-2025-docagent, yang2026alden, fu2026dmap}. 
Despite these scaling improvements, existing pipelines still evaluate retrieved pages as isolated query-image pairs, leaving them highly susceptible to topically overlapping but answer-void distractors. 
To bridge these gaps, \method\ casts validation as a joint, cross-page verification problem and bypasses cascading failure modes via memory-guided iterative reasoning. 
Detailed related works are deferred to Appendix~\ref{app:detailed-related-work}.

\begin{figure*}[t]
  \centering
  \includegraphics[width=\textwidth]{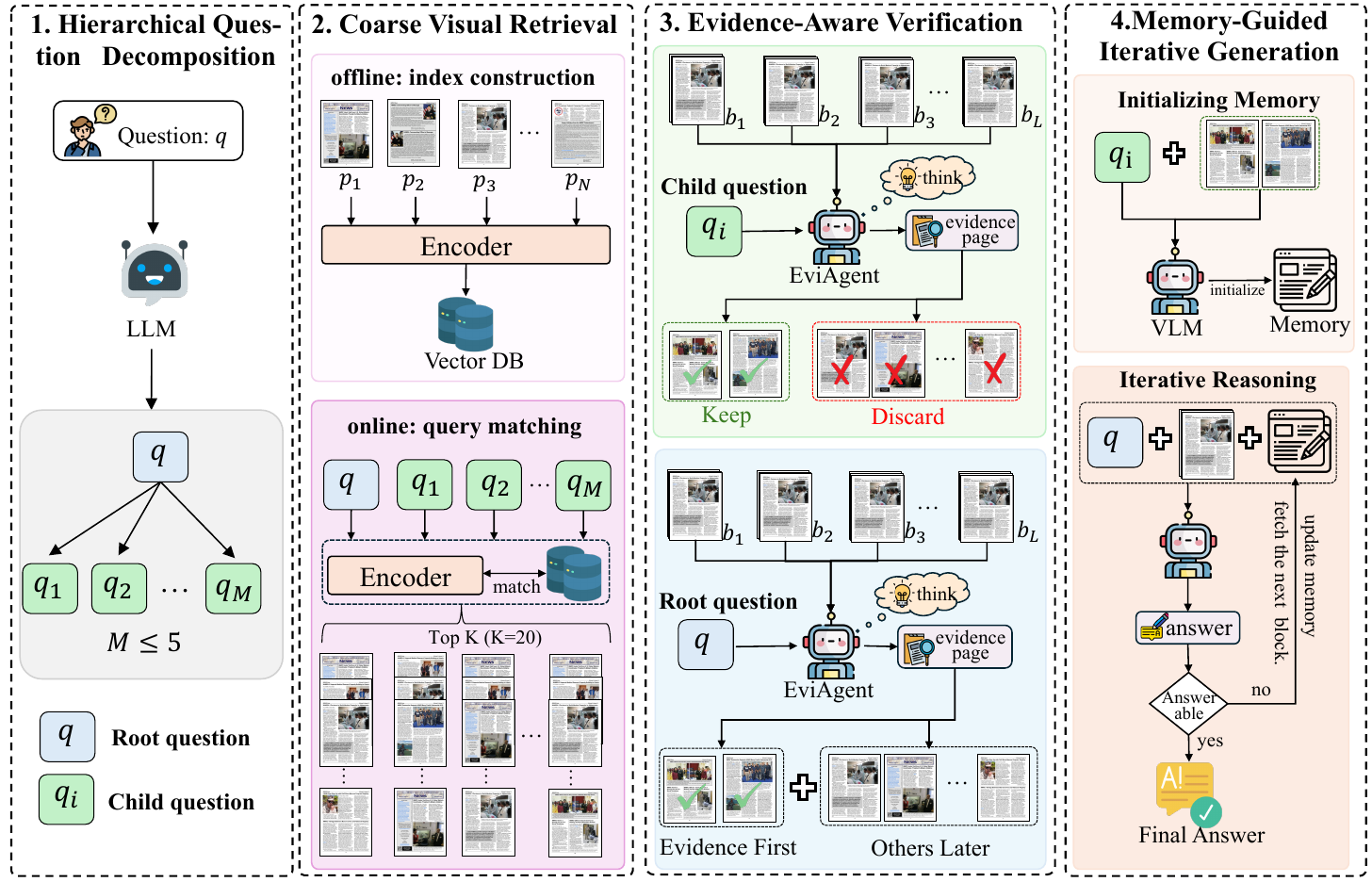}
  \caption{Overview of \method. The system first builds a shallow hierarchy from
  the original question to atomic child questions, retrieves Top-$K$ candidate
  pages for both root and child questions with a visual retriever, verifies and
  reranks grouped-$k$ candidates with \eviagent, and performs memory-guided
  iterative reasoning over the verified evidence order.}
  \label{fig:model-overview}
\end{figure*}

\section{Method}

\subsection{Overview and Task Formulation}
Let a document be represented as an ordered sequence of page images $D=\{p_1, p_2, \ldots, p_N\}$ and $q$ denote a user question in a closed-domain setting. The final answer is assumed to be grounded in $D$, although the supporting evidence may be sparsely distributed across multiple non-contiguous pages. When the document length $N$ scales to hundreds of pages, direct end-to-end VLM ingestion becomes computationally prohibitive due to context window constraints. To bypass this bottleneck, \method\ decomposes long-document QA into four cooperative stages: (1) hierarchical question decomposition, (2) coarse visual page retrieval, (3) evidence-aware page verification, and (4) memory-guided iterative generation. A core design rationale of our framework is the multi-granularity operation across stages: coarse retrieval is strictly recall-oriented, \eviagent\ conducts fine-grained verification over grouped page candidates via cross-page reasoning, and the final generator iteratively consumes the verified evidence sequence backed by an explicit memory. The overall pipeline is illustrated in Figure~\ref{fig:model-overview}.

\subsection{Hierarchical Question Decomposition}

Given a complex root question $q$, this stage constructs a shallow reasoning hierarchy, as illustrated in Figure~\ref{fig:model-overview}(1). Specifically, a lightweight LLM extracts key entities, structural attributes, constraints, and logical relations from $q$ to instantiate a set of atomic child questions:
\begin{equation}
\mathcal{Q}(q)=\{q_1, q_2, \ldots, q_m\}, \qquad m \leq 5.
\end{equation}
Each child question $q_i$ is formulated as a distinct natural query covering a non-redundant subset of the global information need, while strictly avoiding near-paraphrases of the full prompt. 

Crucially, partitioning the global query into localized targets reduces the inherent difficulty of multi-hop reasoning. This upfront decomposition prevents retrieval omissions typical of standalone root queries, ensuring core evidence is captured before downstream generation. The corresponding prompt template and example are detailed in Appendix~\ref{app:decomp-prompt} and ~\ref{app:decomp-case-study}.

\subsection{Coarse Visual Page Retrieval}

Following decomposition, we employ \texttt{Ops-ColQwen3-4B}~\cite{ops_colqwen3_4b}, a ColPali-style late-interaction multimodal retriever~\cite{faysse2025colpali,khattab2020colbert}, to construct a candidate page pool from rendered document images. As illustrated in Figure~\ref{fig:model-overview}(2), this retrieval stage is split into offline index construction and online query matching.

\paragraph{Offline Index Construction.} Each document page image $p_i \in D$ is encoded offline into a multi-vector representation to preserve fine-grained visual features:
\begin{equation}
\mathbf{v}_{p_i} = \mathcal{E}_{\text{page}}(p_i), \qquad i=1,2,\ldots,N.
\end{equation}
The derived page embeddings are subsequently stored to build the global document index.

\paragraph{Online Query Matching.} Upon receiving a root question $q$, the stage online encodes both $q$ and its derived atomic child questions $q_i \in \mathcal{Q}(q)$ to obtain their respective text embeddings:
\begin{equation}
\mathbf{v}_q = \mathcal{E}_{\text{query}}(q), \quad \mathbf{v}_{q_i} = \mathcal{E}_{\text{query}}(q_i).
\end{equation}
These query embeddings are then matched against the offline page index using the late-interaction operator to compute similarity scores across all document pages. Based on these scores, we rank and retain the Top-$K$ ($K=20$) candidate pages for the root question and each child question, denoted as $T_q$ and $T_{q_i}$ respectively.

\subsection{Evidence-Aware Page Verification}
To filter out topically related but answer-void distractors from the coarse retrieval pool, we introduce \eviagent, a specialized model trained to perform binary evidence page verification via cross-page reasoning. As illustrated in Figure~\ref{fig:model-overview}(3), this stage takes the coarse candidate lists $T_q$ and $T_{q_i}$ as input, partitions them into grouped-$k$ sliding windows, and leverages the trained agent to output structured decisions.

\paragraph{Training \eviagent\ with Evidence-Aware GRPO.} Inspired by DocR1~\cite{xiong2025docr1}, we train \eviagent\ using an Evidence-Aware Group Relative Policy Optimization (EviGRPO) pipeline to specialize a VLM for multi-page document verification. The policy is optimized across six open-source multi-page datasets~\cite{van-landeghem2023dude,tito2022mpdocvqa,zhao-etal-2022-multihiertt,schimanski2026pdfqa,yu2026sciegqa,tanaka2023slidevqa} with statistics detailed in Appendix~\ref{app:training_dataset}.

During the training phase, the VLM is constrained to generate a structured content containing three mandatory fields: $o = (\tau, \mathbf{y}, a)$. The corresponding prompt template is detailed in Appendix~\ref{app:eviagent-prompt}.

\begin{itemize}
    \item \textbf{Reasoning Trace} ($\tau$): Contained within the \texttt{<think>} tag, $\tau$ encapsulates intra-page visual inspection and cross-page reasoning.
    \item \textbf{Evidence Decisions} ($\mathbf{y} \in \{T, F\}^k$): Enclosed within the \texttt{<evidence\_page>} tag, $\mathbf{y}$ enforces a strict one-to-one binary sequence matching the exact order of input images to prevent evaluation omissions.
    \item \textbf{Final Answer} ($a$): Wrapped within the \texttt{<answer>} tag, $a$ delivers the terminal text response derived from the verified evidence.
\end{itemize}

The policy is optimized via a weighted joint reward function:
\begin{equation}
R = \lambda_{\text{fmt}} r_{\text{fmt}} + \lambda_{\text{evi}} r_{\text{evi}} + \lambda_{\text{acc}} r_{\text{acc}},
\end{equation}
where $r_{\text{fmt}}$ enforces schema adherence, and $r_{\text{acc}}$ evaluates final answer correctness. The evidence reward $r_{\text{evi}}$ computes the $F_1$ score against ground-truth page labels to accurately measure verification performance and prevent reward hacking.

\paragraph{Fine-Grained Page Validation.} During online inference, each retrieved candidate list is split into $L = \lceil K/k \rceil$ blocks of size $k$: $b_t^k = \{p_{(t-1)k+1}, \ldots, p_{\min(tk,K)}\}$, where $t \in \{1, \ldots, L\}$ is the block index. Each block is verified jointly by the trained \eviagent. To prevent information fragmentation across boundaries, all pages verified as true evidence are aggregated into a final consolidated block for a global cross-block synthesis. The refined signals are then routed to two distinct downstream policies:

\begin{itemize}
    \item \textbf{Discarding for Child Questions}: For each child question $q_i$, we extract only the verified evidence pages:
    \begin{equation}
    E(q_i) = \{p \in T_{q_i} : \mathbf{y}_p = T\}.
    \end{equation}
    Non-evidence pages are discarded to restrict downstream generation strictly to localized, verified facts.
    \item \textbf{Reranking for the Root Question}: For the root question $q$, let $E(q)$ denote the verified evidence page set. We construct a prioritized sequence $\Pi(q)$ while preserving the original retriever ranking order within each partition:
    \begin{equation}
    \Pi(q) = E(q) \oplus (T_q \setminus E(q)),
    \end{equation}
  where $\oplus$ denotes sequence concatenation. Both $E(q)$ and the remaining subset $T_q \setminus E(q)$ strictly inherit the original ranking order determined by the coarse visual retriever. Non-evidence pages are retained as a backup suffix to secure recall for downstream iterative reasoning.
  \end{itemize}

\subsection{Memory-Guided Iterative Generation}

This final stage leverages the resolved child questions and the prioritized sequence $\Pi(q)$ to generate the terminal answer via a two-step, memory-enhanced pipeline. As illustrated in Figure~\ref{fig:model-overview}(4), the process consists of initializing an explicit memory layer with sub-question contexts and iteratively reasoning over the root question.

\paragraph{Initializing Memory with Sub-Questions.} Before addressing the root query, the framework sequentially resolves each child question $q_i \in \mathcal{Q}(q)$ over its verified evidence block $E(q_i)$ using \eviagent:
\begin{equation}
(\tau_i, \mathbf{y}_i, a_i) = \mathcal{A}\big(q_i, E(q_i)\big),
\end{equation}
where $\mathcal{A}$ denotes the generative policy of \eviagent. Initializing the memory as $M_0 = \emptyset$, the state updates cumulatively by appending the complete triplets:
\begin{equation}
M_i = M_{i-1} \cup \{(q_i, \tau_i, a_i)\}.
\end{equation}
This mechanism distills sub-question verification into informative textual facts, seeding the working memory with both mid-level execution traces and terminal local answers.

\paragraph{Iterative Reasoning for the Root Question.} Upon consolidating the sub-question memory $M_m$ (where $m = |\mathcal{Q}(q)|$), \method\ evaluates the root question $q$ by scanning the prioritized sequence $\Pi(q)$ via a sliding window of size $k$. At execution round $t$, \eviagent\ ingests the root query, the persistent memory state, and the $t$-th evidence block $b_t^k \subset \Pi(q)$:
\begin{equation}
(\tau_t, \mathbf{y}_t, a_t) = \mathcal{A}\big(q, b_t^k, M_{m+t-1}\big).
\end{equation}
The execution terminates immediately, returning $a_t$ as the global response if $a_t \neq \alpha_{\emptyset}$, where $\alpha_{\emptyset}$ denotes the designated abstention token (i.e., \texttt{NOT\_ANSWERABLE}). Otherwise, the current reasoning trace $\tau_t$ is appended to the memory to guide the next window:
\begin{equation}
M_{m+t} = M_{m+t-1} \cup \{(q, \tau_t)\}.
\end{equation}
This iterative accumulation continues until a valid answer is produced or $\Pi(q)$ is exhausted. By maintaining text-based memory across rounds, this recurrent design forces the model to synthesize historical context globally, preventing critical information omission inherent in local window constraints.

\section{Experiments}
\begin{table}[h]
  \centering
  \scalebox{0.5}{
  \resizebox{\textwidth}{!}{%
  \begin{tabular}{lcccc}
    \toprule
    Benchmark & \# QA Samples & Min. Pages & Max. Pages & Avg. Pages \\
    \midrule
    PaperTab & 393 & 2 & 152 & 10.7 \\
    FetaTab & 1016 & 2 & 216 & 16.3 \\
    MMLongBench & 1,082 & 9 & 468 & 47.8 \\
    LongDocURL & 2,325 & 51 & 149 & 89.0 \\
    \bottomrule
  \end{tabular}}
  }
  \caption{Benchmark statistics for the four-dataset evaluation protocol. Average
  page counts are rounded to one decimal place.}
  \label{tab:benchmark-details}
\end{table}

\begin{table*}[!t]
  \centering
  \scriptsize
  \resizebox{\textwidth}{!}{%
  \begin{tabular}{llcccc}
    \toprule
    Method & Param. & PaperTab & FetaTab & MMLongBench & LongDocURL \\
    & & (Acc.) & (Acc.) & (Acc.) & (Acc.) \\
    \midrule
    M3DocRAG~\cite{cho2024m3docrag} & 7B & 28.5 & 63.8 & 36.2 & 49.0 \\
    MDocAgent~\cite{han2025mdocagent} & 7B & 30.0 & 66.3 & 38.5 & 46.9 \\
    MoLoRAG+~\cite{wu-etal-2025-molorag} & 7B & \underline{31.0} & \underline{69.2} & 41.0 & 51.9 \\
    ALDEN~\cite{yang2026alden} & 7B & 24.5 & 62.3 & 39.2 & 55.1 \\
    URaG~\cite{shi2026urag} & 7B & -- & -- & 33.8 & 52.2 \\
    Doc-$V^\star$~\cite{zheng2026docvstar} & 7B & -- & -- & \underline{42.1} & \underline{56.3} \\
    \midrule
    \method & 8B & \textbf{44.0} & \textbf{72.9} & \textbf{48.2} & \textbf{65.7} \\
    \bottomrule
  \end{tabular}}
  \caption{Main experimental results across the four benchmarks. 
The best results are highlighted in \textbf{bold}, and the \underline{second-best} results are underlined. 
The symbol ``--'' indicates that the baseline result is not reported. 
ALDEN uses the ColQwen+ColBERT pipeline, and Doc-$V^\star$ refers to the GRPO-optimized model.}
  \label{tab:main-generation}
\end{table*}

\begin{table*}[t]
  \centering
  \scriptsize
  \resizebox{\textwidth}{!}{%
  \begin{tabular}{lccccc}
    \toprule
    Variant & PaperTab & FetaTab & MMLongBench & LongDocURL & Avg. \\
    \midrule
    \method 
    & \textbf{44.0} 
    & \textbf{72.9} 
    & \textbf{48.2} 
    & \textbf{65.7} 
    & \textbf{57.7} \\
    w/o Evidence-Aware Verification 
    & 39.7 {\scriptsize(-4.3)} 
    & 69.6 {\scriptsize(-3.3)} 
    & 44.5 {\scriptsize(-3.7)} 
    & 59.1 {\scriptsize(-6.6)} 
    & 53.2 {\scriptsize(-4.5)} \\
    w/o Hierarchical Decomposition 
    & 41.3 {\scriptsize(-2.7)} 
    & 70.5 {\scriptsize(-2.4)} 
    & 45.8 {\scriptsize(-2.4)} 
    & 62.4 {\scriptsize(-3.3)} 
    & 55.0 {\scriptsize(-2.7)} \\
    w/o Iterative Reasoning 
    & 41.9 {\scriptsize(-2.1)} 
    & 72.0 {\scriptsize(-0.9)} 
    & 46.1 {\scriptsize(-2.1)} 
    & 63.2 {\scriptsize(-2.5)} 
    & 55.8 {\scriptsize(-1.9)} \\
    \bottomrule
  \end{tabular}}
\caption{Ablation analysis of individual architectural modules. 
For \textit{w/o Evidence-Aware Verification}, \method's Step 3 is bypassed; sub-questions in Step 4 initialize memory directly using the unverified Top-$K$ coarse retrieval, and the root question iteratively reasons over the un-reranked sequence. 
  For \textit{w/o Hierarchical Decomposition}, \method's Step 1 and the memory initialization phase in Step 4 are omitted, while the iterative reasoning loop remains active. 
  For \textit{w/o Iterative Reasoning}, the sliding window in \method's Step 4 is deactivated, force-feeding the Top-5 candidate pages into the generator in a single-pass manner.}
\label{tab:generation-ablation}
\end{table*}

\subsection{Experimental Setup}
\label{sec:exp-setup}

\paragraph{Benchmarks.}
We evaluate \method\ on four benchmarks: PaperTab and FetaTab~\cite{hui2024uda}, which target question answering over scientific and Wikipedia-style tables; and MMLongBench~\cite{wang2025mmlongbench} and LongDocURL~\cite{deng-etal-2025-longdocurl}, which feature extended multimodal documents stressing cross-page retrieval, visual grounding, and multi-hop reasoning. Structural statistics are provided in Table~\ref{tab:benchmark-details}.

\paragraph{Implementation Details.} For hierarchical question decomposition, a lightweight \texttt{Qwen3.5-4B} model is deployed by default to generate at most five child questions. The baseline coarse retriever is configured as \texttt{ops-ColQwen3-4B}~\cite{ops_colqwen3_4b}. Both root and child queries retrieve the Top-$K$ ($K=20$) candidate pages, which are subsequently verified using a grouped window size of $k=7$. We instantiate \eviagent\ using \texttt{Qwen3-VL-8B-Instruct}~\cite{qwen2025qwen3vl} as the foundational backbone. For the joint reward function, the balancing scaling weights are explicitly set as $(\lambda_{\text{fmt}}, \lambda_{\text{evi}}, \lambda_{\text{acc}}) = (0.1, 0.5, 0.4)$, prioritizing the evidence verification performance. Training is executed on an $8\times$ NVIDIA A100 GPU cluster.
\paragraph{Metrics.}
For MMLongBench and LongDocURL, we follow their original evaluation protocols, employing standard accuracy and rule-based string matching tailored to various answer types. For PaperTab and FetaTab, we adopt an LLM-as-a-judge paradigm leveraging Qwen3.5-27B as the evaluator to compute binary accuracy (0 or 1) by assessing whether the generated response semantically matches the ground truth. Additionally, retrieval-related performance is quantified via Recall@K, NDCG@K, and MRR@K, with detailed mathematical formulations deferred to Appendix~\ref{app:retrieval-metrics}.

\begin{table*}[t]
  \centering
  \scriptsize
  \resizebox{\textwidth}{!}{%
  \begin{tabular}{lllcccc}
    \toprule
    Method & Generator & Retriever & PaperTab & FetaTab & MMLongBench & LongDocURL \\
    & & & (Acc.) & (Acc.) & (Acc.) & (Acc.) \\
    \midrule
    MoLoRAG+~\cite{wu-etal-2025-molorag} 
    & \texttt{Qwen2.5-VL-7B-Instruct} 
    & \texttt{ColQwen2.5} 
    & 31.0 & 69.2 & 41.0 & 51.9 \\
    \midrule
    \multirow{4}{*}{\method}
    & \texttt{Qwen2.5-VL-7B-Instruct} 
    & \texttt{ColQwen2.5} 
    & 34.2 & 69.1 & 42.6 & 55.4 \\
    
    & \texttt{Qwen2.5-VL-7B-Instruct} 
    & \texttt{Ops-ColQwen3-4B} 
    & 38.3 & 71.4 & 45.0 & 60.2 \\
    
    & \texttt{Qwen3-VL-8B-Instruct} 
    & \texttt{ColQwen2.5} 
    & 36.0 & 70.8 & 43.8 & 57.1 \\
    
    & \texttt{Qwen3-VL-8B-Instruct} 
    & \texttt{Ops-ColQwen3-4B} 
    & \textbf{44.0} & \textbf{72.9} & \textbf{48.2} & \textbf{65.7} \\
    \bottomrule
  \end{tabular}}
  \caption{Sensitivity analysis over generation backbones and visual retrievers. 
  MoLoRAG+ is included as the external baseline. The last row corresponds to the main \method\ configuration in Table~\ref{tab:main-generation}. 
  The remaining \method\ rows are controlled variants that change only the listed generator or retriever while keeping the same evidence-verification and iterative-generation pipeline.}
  \label{tab:backbone-retriever}
\end{table*}

\subsection{Main Results}
Table~\ref{tab:main-generation} presents a performance comparison across the four benchmarks. \method\ consistently outperforms all competitive baselines, validating the synergistic effects of its core architectural components.

On PaperTab and FetaTab, \method\ achieves \textbf{44.0\%} and \textbf{72.9\%} accuracy, yielding absolute improvements of +13.0\% and +3.7\% over the strongest baseline (MoLoRAG+), respectively. Tabular documents typically require exact cell alignment; standard retrievers frequently introduce false positives due to repetitive schema structures. \method\ mitigates this bottleneck by executing fine-grained verification via \eviagent, effectively isolating answer-void distractors.

Similarly, on MMLongBench and LongDocURL, \method\ secures \textbf{48.2\%} and \textbf{65.7\%} accuracy, outperforming Doc-$V^\star$ by +6.1\% and +9.4\%, respectively. These tasks necessitate cross-page retrieval and multi-hop reasoning, where single-pass pipelines easily omit critical information. The consistent performance gains demonstrate the efficacy of our memory-guided iterative generation.

\subsection{Ablation Studies}

\paragraph{Contribution of Individual Modules.} 
To isolate the empirical impact of each architectural phase in \method, we evaluate three ablation variants across all benchmarks (Table~\ref{tab:generation-ablation}). Disabling any single stage yields a consistent performance degradation, validating the synergy of our core components.

Specifically, \textit{w/o Evidence-Aware Verification} drops the average score by -4.5\% absolutely. Without Step 3 verification, answer-void distractors occupy the front of the sequence, misleading the generator into premature early stopping and missing genuine evidence. Furthermore, initializing memory over unverified Top-$K$ inputs introduces catastrophic cascading noise. Meanwhile, \textit{w/o Hierarchical Decomposition} compromises the average score by -2.7\%. This demonstrates that executing complex queries as standalone prompts without \method's Step 1 induces severe retrieval omissions, whereas atomic decomposition guarantees the upfront recall of dispersed evidence. Lastly, \textit{w/o Iterative Reasoning} consistently degrades results by -1.9\% on average. This variant deactivates the sliding window in \method's Step 4 and force-feeds the Top-5 pages in a single pass. The result proves that while verification serves as an effective filter, iterative reasoning acts as an indispensable safety net against information omission under local window constraints.

\paragraph{Backbone and Retriever Sensitivity.} To evaluate the robustness of \method\ under varying underlying architectures, we perform a grid sensitivity analysis by cross-combining different generative backbones and visual retrievers (Table~\ref{tab:backbone-retriever}). 

Specifically, when deploying the identical backbone and retriever configuration as MoLoRAG+ (\texttt{Qwen2.5-VL-7B-Instruct} + \texttt{ColQwen2.5}), \method\ outperforms this baseline in almost all benchmarks. It delivers clear absolute improvements of +3.2\% on PaperTab and +3.5\% on LongDocURL, demonstrating that our primary advantages stem strictly from architectural innovations rather than the scaling of foundational models. Furthermore, scaling up either the retriever to \texttt{Ops-ColQwen3-4B} or the generator to \texttt{Qwen3-VL-8B-Instruct} provides independent performance increments across benchmarks.  The combination of both upgraded components yields the best performance, proving that our framework effectively leverages stronger retrieval and generation capabilities to maximize final accuracy.

\begin{table*}[t]
  \centering
  \scriptsize
  \resizebox{\textwidth}{!}{%
  \begin{tabular}{llcccccc}
    \toprule
    \multirow{2}{*}{Top-$K$} 
    & \multirow{2}{*}{Method} 
    & \multicolumn{3}{c}{MMLongBench} 
    & \multicolumn{3}{c}{LongDocURL} \\
    \cmidrule(lr){3-5} \cmidrule(lr){6-8}
    & & Recall & NDCG & MRR & Recall & NDCG & MRR \\
    \midrule

    \multirow{3}{*}{1}
    & Ops-ColQwen3-4B (baseline) 
    & 54.2 & 70.4 & 70.4 & 53.5 & 73.4 & 73.4 \\
    & + Qwen3-VL-8B-Instruct 
    & 54.2 {\scriptsize $(+0.0)$} 
    & 70.6 {\scriptsize $(+0.2)$} 
    & 70.6 {\scriptsize $(+0.2)$} 
    & 52.9 {\scriptsize $(-0.6)$} 
    & 72.1 {\scriptsize $(-1.3)$} 
    & 72.1 {\scriptsize $(-1.3)$} \\
    & + \eviagent
    & 59.4 {\scriptsize $(+5.2)$} 
    & 78.3 {\scriptsize $(+7.9)$} 
    & 78.3 {\scriptsize $(+7.9)$} 
    & 57.3 {\scriptsize $(+3.8)$} 
    & 78.8 {\scriptsize $(+5.4)$} 
    & 78.8 {\scriptsize $(+5.4)$} \\

    \midrule

    \multirow{3}{*}{3}
    & Ops-ColQwen3-4B (baseline) 
    & 75.2 & 73.8 & 77.2 & 74.8 & 73.7 & 81.1 \\
    & + Qwen3-VL-8B-Instruct 
    & 75.6 {\scriptsize $(+0.4)$} 
    & 74.4 {\scriptsize $(+0.6)$} 
    & 77.6 {\scriptsize $(+0.4)$} 
    & 73.7 {\scriptsize $(-1.1)$} 
    & 72.5 {\scriptsize $(-1.2)$} 
    & 79.8 {\scriptsize $(-1.3)$} \\
    & + \eviagent
    & 80.1 {\scriptsize $(+4.9)$} 
    & 80.6 {\scriptsize $(+6.8)$} 
    & 83.7 {\scriptsize $(+6.5)$} 
    & 77.5 {\scriptsize $(+2.7)$} 
    & 77.3 {\scriptsize $(+3.6)$} 
    & 84.9 {\scriptsize $(+3.8)$} \\

    \midrule

    \multirow{3}{*}{5}
    & Ops-ColQwen3-4B (baseline) 
    & 81.4 & 76.2 & 78.2 & 81.0 & 76.4 & 82.0 \\
    & + Qwen3-VL-8B-Instruct 
    & 81.8 {\scriptsize $(+0.4)$} 
    & 76.6 {\scriptsize $(+0.4)$} 
    & 78.5 {\scriptsize $(+0.3)$} 
    & 80.2 {\scriptsize $(-0.8)$} 
    & 75.4 {\scriptsize $(-1.0)$} 
    & 80.7 {\scriptsize $(-1.3)$} \\
    & + \eviagent
    & 85.2 {\scriptsize $(+3.8)$} 
    & 82.2 {\scriptsize $(+6.0)$} 
    & 84.4 {\scriptsize $(+6.2)$} 
    & 83.2 {\scriptsize $(+2.2)$} 
    & 79.7 {\scriptsize $(+3.3)$} 
    & 85.6 {\scriptsize $(+3.6)$} \\

    \bottomrule
  \end{tabular}}
  \caption{Retrieval performance on MMLongBench and LongDocURL under different Top-$K$ settings. Values in parentheses denote the absolute difference compared with the Ops-ColQwen3-4B baseline under the same Top-$K$ setting.}
  \label{tab:retrieval-analysis}
\end{table*}

\begin{figure*}[t]
  \centering
  \includegraphics[width=\textwidth]{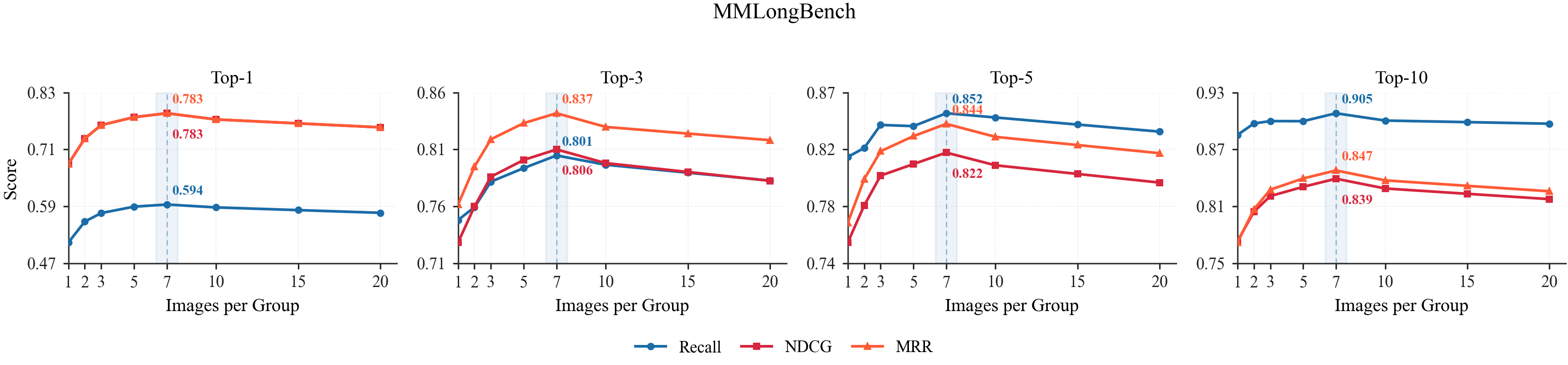}
  \caption{Effect of block size $k$ on MMLongBench retrieval quality.
  A moderate block size provides the best balance between cross-page comparison
  and visual-context overload.}
  \label{fig:retrieval-ablation}
\end{figure*}

\paragraph{Effectiveness of Fine-Grained Page Validation.} To evaluate the impact of the verification stage on retrieval quality, we compare the retrieval metrics of the baseline against the reranked outputs generated by our verifier both before and after training under various Top-$K$ settings on MMLongBench and LongDocURL (Table~\ref{tab:retrieval-analysis}).

Specifically, deploying an untuned \textit{Qwen3-VL-8B-Instruct} yields marginal or even detrimental performance shifts. While it provides negligible fluctuations on MMLongBench (e.g., a modest $+0.2$ point increase in NDCG@1 at Top-1), it consistently degrades retrieval precision on LongDocURL, with NDCG slipping by $-1.3$ points at Top-1 and $-1.0$ points at Top-5. This demonstrates that off-the-shelf VLMs lack the necessary alignment to distinguish authentic evidence from complex multimodal distractors. 
In sharp contrast, our trained \textit{\eviagent} consistently and substantially outperforms the \textit{baseline}. It elevates the NDCG@1 from 70.4\% to 78.3\% on MMLongBench, and from 73.4\% to 78.8\% on LongDocURL. Furthermore, at Top-5, \eviagent\ pushes the final Recall to 85.2\% and 83.2\% on the respective benchmarks. These comprehensive gains across Recall, NDCG, and MRR solidly confirm \eviagent's superior capacity in identifying authentic evidence and filtering out answer-void noise.

\paragraph{Sensitivity to Block Size.} To evaluate the impact of the block size $k$ defined in \method's Step 3, we analyze the retrieval quality under varying scales across MMLongBench (Figure~\ref{fig:retrieval-ablation}). The empirical results indicate that both excessively large and excessively small values of $k$ degrade retrieval quality.

Specifically, a moderate block size (e.g., $k=7$) achieves the optimal balance and yields the highest retrieval metrics. Setting $k$ too small isolates candidate pages. This constrains the model's capacity for cross-page reasoning and comparison. Conversely, expanding $k$ excessively degrades performance. This drop is driven by visual context overload, where the excessive input scale dilutes model attention. Therefore, maintaining a tightly bounded block size $k$ is essential to maximize page validation accuracy.




\section{Conclusion}
In this work, we presented \method, a hierarchical, evidence-driven multimodal RAG framework for closed-domain long-document understanding, whose core innovation lies in its cooperative four-stage pipeline. Specifically, the framework first employs hierarchical question decomposition to partition complex multi-hop queries into atomic child questions, drastically reducing initial retrieval difficulty. Second, coarse visual retrieval leverages rendered page indexing to guarantee upfront evidence recall. Third, fine-grained page verification utilizes the \eviagent\ to perform cross-page reasoning over multi-image blocks, precisely suppressing topically related yet answer-void distractors. Fourth, memory-guided iterative generation accumulates sub-question traces into an explicit text-based memory, dynamically executing sliding-window reasoning to eliminate cascading information omissions. 
Extensive experiments across four benchmarks demonstrate that \method\ significantly outperforms existing open-source baselines. Furthermore, comprehensive ablation studies analyses solidly validate the individual efficacy and systemic synergy of each cooperative phase within our pipeline.

\section*{Limitations}
Despite the state-of-the-art performance achieved by \method, several inherent limitations warrant further investigation. 

First, our framework introduces non-trivial computational overhead during the inference stage. Specifically, the hierarchical question decomposition, the multi-block page verification via the  \eviagent, and the subsequent memory-guided generation inherently mandate multiple sequential LLM/VLM invocations. This compound multi-stage architecture inevitably increases cumulative response latency and inference costs compared to standard single-pass pipelines. 

Second, the memory-guided generation mechanism relies heavily on the quality of textual intermediate reasoning traces; if the initial sub-questions yield severely hallucinated or biased summaries, such errors may cascade into the persistent memory layer and occasionally misguide the terminal root-question synthesis. 

Third, our framework was optimized and evaluated predominantly on born-digital electronic documents with pristine rendering. Its zero-shot generalization capabilities across real-world handwritten inputs, heavily degraded physical scans with text artifacts, or scene-text images captured in natural environments require broader empirical validation in future work.

\section*{Ethical Considerations}

\method\ is designed for closed-domain long-document understanding and may be applied to documents containing sensitive or proprietary information. 
All experiments in this paper are conducted on publicly available benchmarks and open-source datasets. 
We do not collect new user data or perform human-subject annotation. 
Nevertheless, deployment in real-world enterprise or financial scenarios should ensure proper access control, data privacy protection, and auditing of generated answers. 
Because the system may still produce incorrect or unsupported answers when retrieval or evidence verification fails, we recommend using the framework as an assistive tool rather than an autonomous decision-making system in high-stakes domains.

\bibliography{custom}

@inproceedings{lewis2020retrieval,
  title = {Retrieval-Augmented Generation for Knowledge-Intensive {NLP} Tasks},
  author = {Lewis, Patrick and Perez, Ethan and Piktus, Aleksandra and Petroni, Fabio and Karpukhin, Vladimir and Goyal, Naman and K{\"u}ttler, Heinrich and Lewis, Mike and Yih, Wen-tau and Rockt{\"a}schel, Tim and Riedel, Sebastian and Kiela, Douwe},
  booktitle = {Advances in Neural Information Processing Systems 33},
  year = {2020},
  url = {https://papers.nips.cc/paper_files/paper/2020/hash/6b493230205f780e1bc26945df7481e5-Abstract.html}
}

@inproceedings{khattab2020colbert,
  title = {{ColBERT}: Efficient and Effective Passage Search via Contextualized Late Interaction over {BERT}},
  author = {Khattab, Omar and Zaharia, Matei},
  booktitle = {Proceedings of the 43rd International ACM SIGIR Conference on Research and Development in Information Retrieval},
  pages = {39--48},
  year = {2020},
  doi = {10.1145/3397271.3401075}
}

@article{tito2022mpdocvqa,
  title = {Hierarchical Multimodal Transformers for Multi-Page {DocVQA}},
  author = {Tito, Rub{\`e}n and Karatzas, Dimosthenis and Valveny, Ernest},
  journal = {arXiv preprint arXiv:2212.05935},
  year = {2022},
  url = {https://arxiv.org/abs/2212.05935}
}

@inproceedings{zhao-etal-2022-multihiertt,
  title = {{M}ulti{H}iertt: Numerical Reasoning over Multi Hierarchical Tabular and Textual Data},
  author = {Zhao, Yilun and Li, Yunxiang and Li, Chenying and Zhang, Rui},
  booktitle = {Proceedings of the 60th Annual Meeting of the Association for Computational Linguistics (Volume 1: Long Papers)},
  pages = {6588--6600},
  year = {2022},
  address = {Dublin, Ireland},
  publisher = {Association for Computational Linguistics},
  url = {https://aclanthology.org/2022.acl-long.454/},
  doi = {10.18653/v1/2022.acl-long.454}
}

@article{tanaka2023slidevqa,
  title = {{SlideVQA}: A Dataset for Document Visual Question Answering on Multiple Images},
  author = {Tanaka, Ryota and Nishida, Kyosuke and Nishida, Kosuke and Hasegawa, Taku and Saito, Itsumi and Saito, Kuniko},
  journal = {arXiv preprint arXiv:2301.04883},
  year = {2023},
  doi = {10.48550/arXiv.2301.04883},
  url = {https://arxiv.org/abs/2301.04883}
}

@article{van-landeghem2023dude,
  title = {Document Understanding Dataset and Evaluation ({DUDE})},
  author = {Van Landeghem, Jordy and Tito, Rub{\'e}n and Borchmann, {\L}ukasz and Pietruszka, Micha{\l} and J{\'o}ziak, Pawe{\l} and Powalski, Rafa{\l} and Jurkiewicz, Dawid and Coustaty, Micka{\"e}l and Ackaert, Bertrand and Valveny, Ernest and Blaschko, Matthew and Moens, Sien and Stanis{\l}awek, Tomasz},
  journal = {arXiv preprint arXiv:2305.08455},
  year = {2023},
  doi = {10.48550/arXiv.2305.08455},
  url = {https://arxiv.org/abs/2305.08455}
}

@inproceedings{faysse2025colpali,
  title = {{ColPali}: Efficient Document Retrieval with Vision Language Models},
  author = {Faysse, Manuel and Sibille, Hugues and Wu, Tony and Omrani, Bilel and Viaud, Gautier and Hudelot, C{\'e}line and Colombo, Pierre},
  booktitle = {The Thirteenth International Conference on Learning Representations},
  year = {2025},
  url = {https://openreview.net/forum?id=ogjBpZ8uSi}
}

@article{xiong2025docr1,
  title = {{DocR1}: Evidence Page-Guided {GRPO} for Multi-Page Document Understanding},
  author = {Xiong, Junyu and Wang, Yonghui and Zhao, Weichao and Liu, Chenyu and Yin, Bing and Zhou, Wengang and Li, Houqiang},
  journal = {arXiv preprint arXiv:2508.07313},
  year = {2025},
  doi = {10.48550/arXiv.2508.07313},
  url = {https://arxiv.org/abs/2508.07313}
}

@inproceedings{chia-etal-2025-longdoc,
  title = {{M}-{L}ong{D}oc: A Benchmark For Multimodal Super-Long Document Understanding And A Retrieval-Aware Tuning Framework},
  author = {Chia, Yew Ken and Cheng, Liying and Chan, Hou Pong and Song, Maojia and Liu, Chaoqun and Aljunied, Mahani and Poria, Soujanya and Bing, Lidong},
  booktitle = {Proceedings of the 2025 Conference on Empirical Methods in Natural Language Processing},
  pages = {9233--9250},
  year = {2025},
  address = {Suzhou, China},
  publisher = {Association for Computational Linguistics},
  url = {https://aclanthology.org/2025.emnlp-main.469/},
  doi = {10.18653/v1/2025.emnlp-main.469}
}

@inproceedings{hui2024uda,
  title = {{UDA}: A Benchmark Suite for Retrieval Augmented Generation in Real-World Document Analysis},
  author = {Hui, Yulong and Lu, Yao and Zhang, Huanchen},
  booktitle = {Advances in Neural Information Processing Systems 37: Datasets and Benchmarks Track},
  year = {2024},
  doi = {10.52202/079017-2145},
  url = {https://proceedings.neurips.cc/paper_files/paper/2024/hash/7c06759d1a8567f087b02e8589454917-Abstract-Datasets_and_Benchmarks_Track.html}
}

@article{cho2024m3docrag,
  title = {{M3DocRAG}: Multi-modal Retrieval is What You Need for Multi-page Multi-document Understanding},
  author = {Cho, Jaemin and Mahata, Debanjan and Irsoy, Ozan and He, Yujie and Bansal, Mohit},
  journal = {arXiv preprint arXiv:2411.04952},
  year = {2024},
  url = {https://arxiv.org/abs/2411.04952}
}

@article{han2025mdocagent,
  title = {{MDocAgent}: A Multi-Modal Multi-Agent Framework for Document Understanding},
  author = {Han, Siwei and Xia, Peng and Zhang, Ruiyi and Sun, Tong and Li, Yun and Zhu, Hongtu and Yao, Huaxiu},
  journal = {arXiv preprint arXiv:2503.13964},
  year = {2025},
  url = {https://arxiv.org/abs/2503.13964}
}

@inproceedings{jain-etal-2025-simpledoc,
  title = {{SimpleDoc}: Multi-Modal Document Understanding with Dual-Cue Page Retrieval and Iterative Refinement},
  author = {Jain, Chelsi and Wu, Yiran and Zeng, Yifan and Liu, Jiale and Dai, Shengyu and Shao, Zhenwen and Wu, Qingyun and Wang, Huazheng},
  booktitle = {Proceedings of the 2025 Conference on Empirical Methods in Natural Language Processing},
  pages = {28410--28427},
  year = {2025},
  address = {Suzhou, China},
  publisher = {Association for Computational Linguistics},
  url = {https://aclanthology.org/2025.emnlp-main.1443/},
  doi = {10.18653/v1/2025.emnlp-main.1443}
}

@inproceedings{sun-etal-2025-docagent,
  title = {{D}oc{A}gent: An Agentic Framework for Multi-Modal Long-Context Document Understanding},
  author = {Sun, Li and He, Liu and Jia, Shuyue and He, Yangfan and You, Chenyu},
  booktitle = {Proceedings of the 2025 Conference on Empirical Methods in Natural Language Processing},
  pages = {17701--17716},
  year = {2025},
  address = {Suzhou, China},
  publisher = {Association for Computational Linguistics},
  url = {https://aclanthology.org/2025.emnlp-main.893/},
  doi = {10.18653/v1/2025.emnlp-main.893}
}

@article{schimanski2026pdfqa,
  title = {pdf{QA}: Diverse, Challenging, and Realistic Question Answering over {PDF}s},
  author = {Schimanski, Tobias and Kolli, Imene and Fan, Yu and Vaghefi, Ario Saeid and Ni, Jingwei and Ash, Elliott and Leippold, Markus},
  journal = {arXiv preprint arXiv:2601.02285},
  year = {2026},
  doi = {10.48550/arXiv.2601.02285},
  url = {https://arxiv.org/abs/2601.02285}
}

@article{yu2026sciegqa,
  title = {{SciEGQA}: A Dataset for Scientific Evidence-Grounded Question Answering and Reasoning},
  author = {Yu, Wenhan and Zhang, Zhaoxi and Chen, Wang and Qi, Guanqiang and Li, Weikang and Sha, Lei and Xia, Deguo and Huang, Jizhou},
  journal = {arXiv preprint arXiv:2511.15090},
  year = {2025},
  doi = {10.48550/arXiv.2511.15090},
  url = {https://arxiv.org/abs/2511.15090}
}

@article{wang2025mmlongbench,
  title = {{MMLongBench}: Benchmarking Long-Context Vision-Language Models Effectively and Thoroughly},
  author = {Wang, Zhaowei and Yu, Wenhao and Ren, Xiyu and Zhang, Jipeng and Zhao, Yu and Saxena, Rohit and Cheng, Liang and Wong, Ginny and See, Simon and Minervini, Pasquale and Song, Yangqiu and Steedman, Mark},
  journal = {arXiv preprint arXiv:2505.10610},
  year = {2025},
  url = {https://arxiv.org/abs/2505.10610}
}

@inproceedings{deng-etal-2025-longdocurl,
  title = {{LongDocURL}: a Comprehensive Multimodal Long Document Benchmark Integrating Understanding, Reasoning, and Locating},
  author = {Deng, Chao and Yuan, Jiale and Bu, Pi and Wang, Peijie and Li, Zhong-Zhi and Xu, Jian and Li, Xiao-Hui and Gao, Yuan and Song, Jun and Zheng, Bo and Liu, Cheng-Lin},
  booktitle = {Proceedings of the 63rd Annual Meeting of the Association for Computational Linguistics (Volume 1: Long Papers)},
  pages = {1135--1159},
  year = {2025},
  address = {Vienna, Austria},
  publisher = {Association for Computational Linguistics},
  url = {https://aclanthology.org/2025.acl-long.57/},
  doi = {10.18653/v1/2025.acl-long.57}
}

@inproceedings{wu-etal-2025-molorag,
  title = {{MoLoRAG}: Bootstrapping Document Understanding via Multi-modal Logic-aware Retrieval},
  author = {Wu, Xixi and Tan, Yanchao and Hou, Nan and Zhang, Ruiyang and Cheng, Hong},
  booktitle = {Proceedings of the 2025 Conference on Empirical Methods in Natural Language Processing},
  pages = {14024--14045},
  year = {2025},
  address = {Suzhou, China},
  publisher = {Association for Computational Linguistics},
  url = {https://aclanthology.org/2025.emnlp-main.708/},
  doi = {10.18653/v1/2025.emnlp-main.708}
}

@inproceedings{zhao-etal-2024-tapera,
  title = {{TaPERA}: Enhancing Faithfulness and Interpretability in Long-Form Table {QA} by Content Planning and Execution-based Reasoning},
  author = {Zhao, Yilun and Chen, Lyuhao and Cohan, Arman and Zhao, Chen},
  booktitle = {Proceedings of the 62nd Annual Meeting of the Association for Computational Linguistics (Volume 1: Long Papers)},
  pages = {12824--12840},
  year = {2024},
  address = {Bangkok, Thailand},
  publisher = {Association for Computational Linguistics},
  url = {https://aclanthology.org/2024.acl-long.692/},
  doi = {10.18653/v1/2024.acl-long.692}
}

@inproceedings{suri-etal-2025-visdom,
  title = {{V}is{D}o{M}: Multi-Document {QA} with Visually Rich Elements Using Multimodal Retrieval-Augmented Generation},
  author = {Suri, Manan and Mathur, Puneet and Dernoncourt, Franck and Goswami, Kanika and Rossi, Ryan A. and Manocha, Dinesh},
  booktitle = {Proceedings of the 2025 Conference of the Nations of the Americas Chapter of the Association for Computational Linguistics: Human Language Technologies (Volume 1: Long Papers)},
  pages = {6088--6109},
  year = {2025},
  address = {Albuquerque, New Mexico},
  publisher = {Association for Computational Linguistics},
  url = {https://aclanthology.org/2025.naacl-long.310/},
  doi = {10.18653/v1/2025.naacl-long.310}
}

@article{zheng2026docvstar,
  title = {{Doc-V*}: Coarse-to-Fine Interactive Visual Reasoning for Multi-Page Document {VQA}},
  author = {Zheng, Yuanlei and Fu, Pei and Li, Hang and Wang, Ziyang and Zhang, Yuyi and Ruan, Wenyu and Zhang, Xiaojin and Wei, Zhongyu and Luo, Zhenbo and Luan, Jian and Chen, Wei and Bai, Xiang},
  journal = {arXiv preprint arXiv:2604.13731},
  year = {2026},
  doi = {10.48550/arXiv.2604.13731},
  url = {https://arxiv.org/abs/2604.13731}
}

@inproceedings{chen-etal-2025-vlm,
  title = {{VLM} Is a Strong Reranker: Advancing Multimodal Retrieval-augmented Generation via Knowledge-enhanced Reranking and Noise-injected Training},
  author = {Chen, Zhanpeng and Xu, Chengjin and Qi, Yiyan and Jiang, Xuhui and Guo, Jian},
  booktitle = {Findings of the Association for Computational Linguistics: EMNLP 2025},
  pages = {8140--8158},
  year = {2025},
  address = {Suzhou, China},
  publisher = {Association for Computational Linguistics},
  url = {https://aclanthology.org/2025.findings-emnlp.432/},
  doi = {10.18653/v1/2025.findings-emnlp.432}
}

@article{xu2025mmr5,
  title = {{MM-R5}: MultiModal Reasoning-Enhanced ReRanker via Reinforcement Learning for Document Retrieval},
  author = {Xu, Mingjun and Dong, Jinhan and Hou, Jue and Wang, Zehui and Li, Sihang and Gao, Zhifeng and Zhong, Renxin and Cai, Hengxing},
  journal = {arXiv preprint arXiv:2506.12364},
  year = {2025},
  doi = {10.48550/arXiv.2506.12364},
  url = {https://arxiv.org/abs/2506.12364}
}

@inproceedings{fu2026dmap,
  title = {{DMAP}: Human-Aligned Structural Document Map for Multimodal Document Understanding},
  author = {Fu, ShunLiang and Zhang, Yanxin and Xiang, Yixin and Du, Xiaoyu and Tang, Jinhui},
  booktitle = {Proceedings of the ACM Web Conference 2026},
  year = {2026},
  url = {https://arxiv.org/abs/2601.18203}
}

@article{yang2026alden,
  title = {{ALDEN}: Reinforcement Learning for Active Navigation and Evidence Gathering in Long Documents},
  author = {Yang, Tianyu and Ruas, Terry and Tian, Yijun and Wahle, Jan Philip and Kurzawe, Daniel and Gipp, Bela},
  journal = {arXiv preprint arXiv:2510.25668},
  year = {2026},
  url = {https://arxiv.org/abs/2510.25668}
}

@article{shi2026urag,
  title = {{URaG}: Unified Retrieval and Generation for Multimodal Document Understanding},
  author = {Shi, Yaya and Zhang, Jingyun and Liu, Yuliang and Li, Hongliang and Bai, Xiang},
  journal = {arXiv preprint arXiv:2603.12189},
  year = {2026},
  url = {https://arxiv.org/abs/2603.12189}
}

@inproceedings{tang2023udop,
  title = {{UDOP}: Unified Document Processing with Vision, Text and Layout},
  author = {Tang, Zineng and Yang, Ziyi and Wang, Guoxin and Fang, Yuwei and Liu, Yang and Zhu, Chenguang and Zeng, Michael and Zhang, Cha and Bansal, Mohit},
  booktitle = {Proceedings of the IEEE/CVF Conference on Computer Vision and Pattern Recognition},
  pages = {24423--24433},
  year = {2023},
  url = {https://openaccess.thecvf.com/content/CVPR2023/html/Tang_UDOP_Unified_Document_Processing_With_Vision_Text_and_Layout_CVPR_2023_paper.html}
}

@inproceedings{appalaraju2024docformerv2,
  title = {{DocFormerv2}: Local Features for Document Understanding},
  author = {Appalaraju, Srikar and Jasani, Bhavan and Kota, Bhargava Urala and Xie, Yusheng and Manmatha, R.},
  booktitle = {Proceedings of the AAAI Conference on Artificial Intelligence},
  year = {2024},
  url = {https://ojs.aaai.org/index.php/AAAI/article/view/27887}
}

@inproceedings{ye2023ureader,
  title = {{UReader}: Universal {OCR}-Free Visually-Situated Language Understanding with Multimodal Large Language Model},
  author = {Ye, Jiabo and Hu, Anwen and Xu, Haiyang and Ye, Qinghao and Yan, Ming and Dan, Yaya and Zhao, Chenliang and Xu, Guohai and Li, Chen and Tian, Junfeng and Qian, Qi and Zhang, Ji and Huang, Fei},
  booktitle = {Findings of the Association for Computational Linguistics: EMNLP 2023},
  pages = {2841--2858},
  year = {2023},
  url = {https://aclanthology.org/2023.findings-emnlp.187/},
  doi = {10.18653/v1/2023.findings-emnlp.187}
}

@article{ye2023mplugdocowl,
  title = {{mPLUG-DocOwl}: Modularized Multimodal Large Language Model for Document Understanding},
  author = {Ye, Jiabo and Hu, Anwen and Xu, Haiyang and Ye, Qinghao and Yan, Ming and Xu, Guohai and Li, Chen and Tian, Junfeng and Qian, Qi and Zhang, Ji and Huang, Fei},
  journal = {arXiv preprint arXiv:2307.02499},
  year = {2023},
  url = {https://arxiv.org/abs/2307.02499}
}

@article{hu2024docowl15,
  title = {{mPLUG-DocOwl} 1.5: Unified Structure Learning for {OCR}-Free Document Understanding},
  author = {Hu, Anwen and Xu, Haiyang and Ye, Jiabo and Yan, Ming and Zhang, Liang and Zhang, Bo and Li, Chen and Zhang, Ji and Jin, Qin and Huang, Fei},
  journal = {arXiv preprint arXiv:2403.12895},
  year = {2024},
  url = {https://arxiv.org/abs/2403.12895}
}

@article{hu2024docowl2,
  title = {{mPLUG-DocOwl2}: High-resolution Compressing for {OCR}-free Multi-page Document Understanding},
  author = {Hu, Anwen and Xu, Haiyang and Ye, Jiabo and Yan, Ming and Zhang, Liang and Zhang, Bo and Li, Chen and Zhang, Ji and Jin, Qin and Huang, Fei},
  journal = {arXiv preprint arXiv:2409.03420},
  year = {2024},
  url = {https://arxiv.org/abs/2409.03420}
}

@article{liu2024textmonkey,
  title = {{TextMonkey}: An {OCR}-Free Large Multimodal Model for Understanding Document},
  author = {Liu, Yuliang and Li, Zhang and Yang, Biao and Li, Chunyuan and Yin, Xu-Cheng and Liu, Cheng-Lin and Jin, Lianwen and Bai, Xiang},
  journal = {arXiv preprint arXiv:2403.04473},
  year = {2024},
  url = {https://arxiv.org/abs/2403.04473}
}

@article{wang2024docllm,
  title = {{DocLLM}: A Layout-Aware Generative Language Model for Multimodal Document Understanding},
  author = {Wang, Dongsheng and Raman, Natraj and Sibue, Mathieu and Ma, Zhiqiang and Babkin, Petr and Kaur, Simerjot and Pei, Yulong and Nourbakhsh, Armineh and Liu, Xiaomo},
  journal = {arXiv preprint arXiv:2401.00908},
  year = {2024},
  doi = {10.48550/arXiv.2401.00908},
  url = {https://arxiv.org/abs/2401.00908}
}

@inproceedings{luo2024layoutllm,
  title = {{LayoutLLM}: Layout Instruction Tuning with Large Language Models for Document Understanding},
  author = {Luo, Chuwei and Shen, Yufan and Zhu, Zhaoqing and Zheng, Qi and Yu, Zhi and Yao, Cong},
  booktitle = {Proceedings of the IEEE/CVF Conference on Computer Vision and Pattern Recognition},
  year = {2024},
  url = {https://arxiv.org/abs/2404.05225}
}

@inproceedings{zhu2025laytokenllm,
  title = {A Simple yet Effective Layout Token in Large Language Models for Document Understanding},
  author = {Zhu, Zhaoqing and Luo, Chuwei and Shao, Zirui and Gao, Feiyu and Xing, Hangdi and Zheng, Qi and Zhang, Ji},
  booktitle = {Proceedings of the IEEE/CVF Conference on Computer Vision and Pattern Recognition},
  year = {2025},
  url = {https://arxiv.org/abs/2503.18434}
}

@article{lv2023kosmos25,
  title = {{KOSMOS}-2.5: A Multimodal Literate Model},
  author = {Lv, Tengchao and Huang, Yupan and Chen, Jingye and Zhao, Yuzhong and Jia, Yilin and Cui, Lei and Ma, Shuming and Chang, Yaoyao and Huang, Shaohan and Wang, Wenhui and Dong, Li and Luo, Weiyao and Wu, Shaoxiang and Wang, Guoxin and Zhang, Cha and Wei, Furu},
  journal = {arXiv preprint arXiv:2309.11419},
  year = {2023},
  doi = {10.48550/arXiv.2309.11419},
  url = {https://arxiv.org/abs/2309.11419}
}

@article{feng2023docpedia,
  title = {{DocPedia}: Unleashing the Power of Large Multimodal Model in the Frequency Domain for Versatile Document Understanding},
  author = {Feng, Hao and Liu, Qi and Liu, Hao and Tang, Jingqun and Zhou, Wengang and Li, Houqiang and Huang, Can},
  journal = {arXiv preprint arXiv:2311.11810},
  year = {2023},
  doi = {10.48550/arXiv.2311.11810},
  url = {https://arxiv.org/abs/2311.11810}
}

@article{wei2024got,
  title = {General {OCR} Theory: Towards {OCR}-2.0 via a Unified End-to-end Model},
  author = {Wei, Haoran and Liu, Chenglong and Chen, Jinyue and Wang, Jia and Kong, Lingyu and Xu, Yanming and Ge, Zheng and Zhao, Liang and Sun, Jianjian and Peng, Yuang and Han, Chunrui and Zhang, Xiangyu},
  journal = {arXiv preprint arXiv:2409.01704},
  year = {2024},
  doi = {10.48550/arXiv.2409.01704},
  url = {https://arxiv.org/abs/2409.01704}
}

@article{mohammadshirazi2024dlava,
  title = {{DLaVA}: Document Language and Vision Assistant for Answer Localization with Enhanced Interpretability and Trustworthiness},
  author = {Mohammadshirazi, Ahmad and Guha Neogi, Pinaki Prasad and Lim, Ser-Nam and Ramnath, Rajiv},
  journal = {arXiv preprint arXiv:2412.00151},
  year = {2024},
  doi = {10.48550/arXiv.2412.00151},
  url = {https://arxiv.org/abs/2412.00151}
}

@inproceedings{nacson2025docvlm,
  title = {{DocVLM}: Make Your {VLM} an Efficient Reader},
  author = {Nacson, Mor Shpigel and Aberdam, Aviad and Ganz, Roy and Ben Avraham, Elad and Golts, Alona and Kittenplon, Yair and Mazor, Shai and Litman, Ron},
  booktitle = {Proceedings of the IEEE/CVF Conference on Computer Vision and Pattern Recognition},
  pages = {29005--29015},
  year = {2025},
  url = {https://openaccess.thecvf.com/content/CVPR2025/html/Nacson_DocVLM_Make_Your_VLM_an_Efficient_Reader_CVPR_2025_paper.html}
}

@inproceedings{wang2025marten,
  title = {Marten: Visual Question Answering with Mask Generation for Multi-modal Document Understanding},
  author = {Wang, Zining and Guan, Tongkun and Fu, Pei and Duan, Chen and Jiang, Qianyi and Guo, Zhentao and Guo, Shan and Luo, Junfeng and Shen, Wei and Yang, Xiaokang},
  booktitle = {Proceedings of the IEEE/CVF Conference on Computer Vision and Pattern Recognition},
  year = {2025},
  url = {https://arxiv.org/abs/2503.14140}
}

@article{yu2025mact,
  title = {Visual Document Understanding and Reasoning: A Multi-Agent Collaboration Framework with Agent-Wise Adaptive Test-Time Scaling},
  author = {Yu, Xinlei and Xu, Chengming and Chen, Zhangquan and Zhang, Yudong and Lu, Shilin and Yang, Cheng and Zhang, Jiangning and Yan, Shuicheng and Hu, Xiaobin},
  journal = {arXiv preprint arXiv:2508.03404},
  year = {2025},
  doi = {10.48550/arXiv.2508.03404},
  url = {https://arxiv.org/abs/2508.03404}
}

@article{shao2024deepseekmath,
    title = {{DeepSeekMath}: Pushing the Limits of Mathematical Reasoning in Open Language Models},
    author = {Shao, Zhihong and Wang, Peiyi and Zhu, Qihao and Xu, Runxin and Song, Junxiao and Bi,
  Xiao and Zhang, Haowei and Zhang, Mingchuan and Li, Y. K. and Wu, Y. and Guo, Daya},
    journal = {arXiv preprint arXiv:2402.03300},
    year = {2024},
    doi = {10.48550/arXiv.2402.03300},
    url = {https://arxiv.org/abs/2402.03300}
  }

@article{qwen2025qwen3vl,
  title = {{Qwen3-VL} Technical Report},
  author = {{Qwen Team}},
  journal = {arXiv preprint arXiv:2511.21631},
  year = {2025},
  doi = {10.48550/arXiv.2511.21631},
  url = {https://arxiv.org/abs/2511.21631}
}

@misc{ops_colqwen3_4b,
  author       = {{OpenSearch-AI}},
  title        = {{Ops-Colqwen3: State-of-the-Art Multimodal Embedding Model for Visual Document Retrieval}},
  year         = {2026},
  howpublished = {\url{https://huggingface.co/OpenSearch-AI/Ops-Colqwen3-4B}},
}

@inproceedings{dong2024encoding,
  title={Encoding spreadsheets for large language models},
  author={Dong, Haoyu and Zhao, Jianbo and Tian, Yuzhang and Xiong, Junyu and Zhou, Mengyu and Lin, Yun and Cambronero, Jos{\'e} and He, Yeye and Han, Shi and Zhang, Dongmei},
  booktitle={Proceedings of the 2024 Conference on Empirical Methods in Natural Language Processing},
  pages={20728--20748},
  year={2024}
}

@inproceedings{xia2024vision,
  title={Vision language models for spreadsheet understanding: Challenges and opportunities},
  author={Xia, Shiyu and Xiong, Junyu and Dong, Haoyu and Zhao, Jianbo and Tian, Yuzhang and Zhou, Mengyu and He, Yeye and Han, Shi and Zhang, Dongmei},
  booktitle={Proceedings of the 3rd Workshop on Advances in Language and Vision Research (ALVR)},
  pages={116--128},
  year={2024}
}

@article{xiong2026priorzero,
  title={PriorZero: Bridging Language Priors and World Models for Decision Making},
  author={Xiong, Junyu and Pu, Yuan and Tang, Jia and Niu, Yazhe},
  journal={arXiv preprint arXiv:2605.12289},
  year={2026}
}

@article{wang2023detect,
  title={Detect any shadow: Segment anything for video shadow detection},
  author={Wang, Yonghui and Zhou, Wengang and Mao, Yunyao and Li, Houqiang},
  journal={IEEE Transactions on Circuits and Systems for Video Technology},
  volume={34},
  number={5},
  pages={3782--3794},
  year={2023},
  publisher={IEEE}
}

@article{wang2023towards,
  title={Towards improving document understanding: An exploration on text-grounding via mllms},
  author={Wang, Yonghui and Zhou, Wengang and Feng, Hao and Zhou, Keyi and Li, Houqiang},
  journal={arXiv preprint arXiv:2311.13194},
  year={2023}
}

@inproceedings{wang2022udoc,
  title={Udoc-gan: Unpaired document illumination correction with background light prior},
  author={Wang, Yonghui and Zhou, Wengang and Lu, Zhenbo and Li, Houqiang},
  booktitle={Proceedings of the 30th ACM International Conference on Multimedia},
  pages={5074--5082},
  year={2022}
}

@article{wang2024root,
  title={Root: Vlm based system for indoor scene understanding and beyond},
  author={Wang, Yonghui and Chen, Shi-Yong and Zhou, Zhenxing and Li, Siyi and Li, Haoran and Zhou, Wengang and Li, Houqiang},
  journal={arXiv preprint arXiv:2411.15714},
  year={2024}
}

@article{wang2024swinshadow,
  title={Swinshadow: Shifted window for ambiguous adjacent shadow detection},
  author={Wang, Yonghui and Liu, Shaokai and Li, Li and Zhou, Wengang and Li, Houqiang},
  journal={ACM Transactions on Multimedia Computing, Communications and Applications},
  volume={20},
  number={11},
  pages={1--20},
  year={2024},
  publisher={ACM New York, NY}
}

\appendix

\section{Detailed Related Work}
\label{app:detailed-related-work}

\subsection{Document Visual Question Answering}
Recent advancements in DocVQA primarily diverge into OCR-based and OCR-free paradigms~\cite{wang2022udoc,wang2023detect,wang2024docllm,wang2024root,wang2024swinshadow,wang2023towards}. OCR-based methods serialize document pages into structured textual representations to serve as inputs for downstream reasoning. Representative architectures leverage multimodal pretraining or specialized attention mechanisms to fuse textual, visual, and layout features, as exemplified by UDOP~\cite{tang2023udop}, DocFormerv2~\cite{appalaraju2024docformerv2}, and DocLLM~\cite{wang2024docllm}. To enhance efficiency and structured generation, LayoutLLM~\cite{luo2024layoutllm} frames layout-sensitive reasoning as an instruction-tuning problem, whereas LayTokenLLM~\cite{zhu2025laytokenllm} employs compact layout tokens to represent spatial bounding boxes without expanding lengthy coordinate sequences. For structured data, TaPERA~\cite{zhao-etal-2024-tapera} incorporates content planning and execution-based reasoning to improve faithfulness in long-form table QA. Although highly effective when structural parsing is reliable, the performance of these methods remains intrinsically upper-bounded by cascading parsing errors and the inevitable loss of visually-situated evidence during text serialization.

OCR-free methods bypass external parsing pipelines by training VLMs to comprehend document images end-to-end. Early foundational architectures prioritized enhancing reading capabilities under constrained token budgets; within this cohort, UReader~\cite{ye2023ureader} introduces auxiliary reading and key-point generation tasks, KOSMOS-2.5~\cite{lv2023kosmos25} pretrains on text-intensive images for structured text generation, DocPedia~\cite{feng2023docpedia} processes visual inputs in the frequency domain, TextMonkey~\cite{liu2024textmonkey} incorporates high-resolution modeling with token filtering, and the mPLUG-DocOwl series~\cite{ye2023mplugdocowl, hu2024docowl15, hu2024docowl2} employs feature compression to scale from single-page to multi-page document understanding. More recently, frameworks have evolved toward specialized reasoning and optimization protocols. Specifically, GOT~\cite{wei2024got} frames vision-based reading as a unified OCR-free generator, whereas DLaVA~\cite{mohammadshirazi2024dlava} and Marten~\cite{wang2025marten} implement visual answer localization and mask prediction to enhance model interpretability. To boost efficiency, DocVLM~\cite{nacson2025docvlm} uses compact textual queries to lower visual token costs, while MACT~\cite{yu2025mact} orchestrates collaborative multi-agent reasoning at test time. To shift from passive reading to active policy optimization, DocR1~\cite{xiong2025docr1} and Doc-$V^\star$~\cite{zheng2026docvstar} leverage evidence-guided GRPO to explicitly train visual reasoning trajectories for multi-page environments.

\subsection{Retrieval-Augmented Generation}
RAG methodologies have evolved from traditional text passage retrieval to multi-modal and structure-aware document discovery. Following the text-centric foundations laid by passage retrieval~\cite{lewis2020retrieval} and token-level late interaction~\cite{khattab2020colbert}, visual RAG systems have increasingly focused on page-level document interfaces. ColPali~\cite{faysse2025colpali} extends late interaction to rendered page images, while M3DocRAG~\cite{cho2024m3docrag} establishes this approach as a core interface for multi-page QA. To decouple retrieval from reading, MDocAgent~\cite{han2025mdocagent} and SimpleDoc~\cite{jain-etal-2025-simpledoc} utilize multi-module collaboration and iterative cascading, while RagVL~\cite{chen-etal-2025-vlm} and MM-R5~\cite{xu2025mmr5} employ instruction-tuning or reinforcement learning to train specialized visual rerankers. Concurrently, efforts have been dedicated to structured or active navigation to handle complex layouts: MoLoRAG~\cite{wu-etal-2025-molorag} leverages page topology graphs for logic-aware traversal, DocAgent~\cite{sun-etal-2025-docagent} and ALDEN~\cite{yang2026alden} introduce memory feedback and active exploration protocols, and DMAP~\cite{fu2026dmap} constructs human-aligned structural maps to guide global understanding. 

\section{Prompt Templates}
\label{app:prompts}
\subsection{Prompt For Hierarchical Question Decomposition}
\label{app:decomp-prompt}
\begin{figure*}[h]
  \centering
  \includegraphics[width=\textwidth,height=0.62\textheight,keepaspectratio]{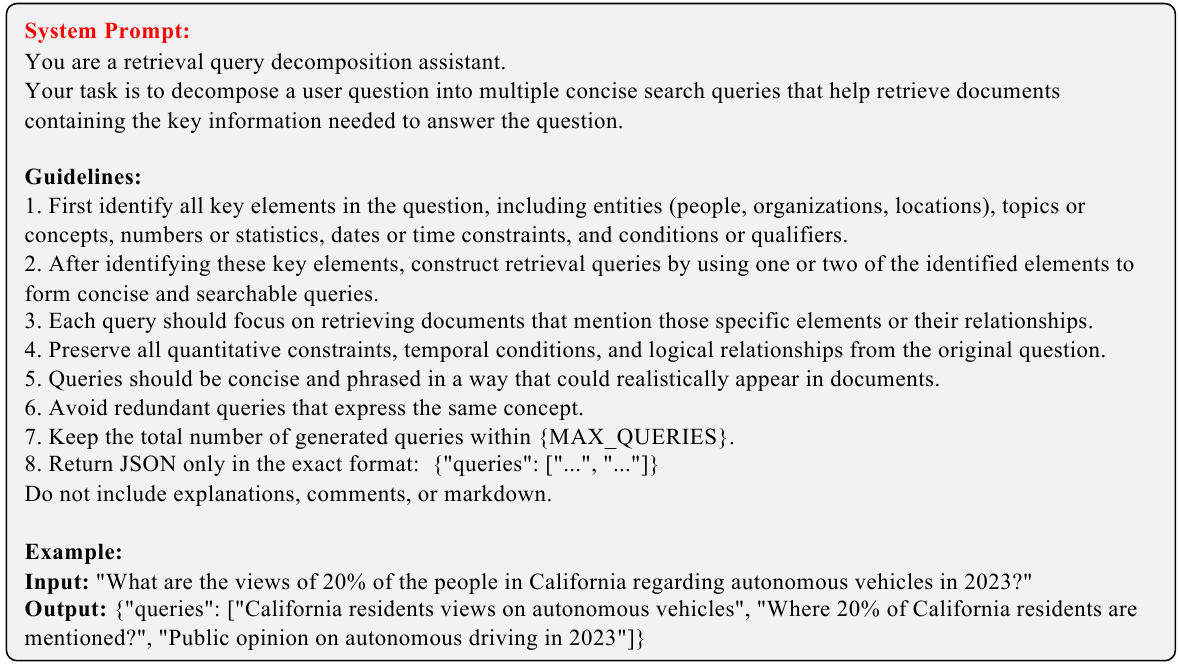}
  \caption{System prompt for hierarchical question decomposition.}
  \label{fig:decomp-prompt}
\end{figure*}

The system prompt for hierarchical question decomposition is detailed in Figure~\ref{fig:decomp-prompt}. Given the root question as input, the model is strictly constrained to output a maximum of five atomic child questions, each formatted strictly as an interrogative sentence.

\subsection{Prompt For EviAgent}
\label{app:eviagent-prompt}
\begin{figure*}[h]
  \centering
  \includegraphics[width=\textwidth,height=0.62\textheight,keepaspectratio]{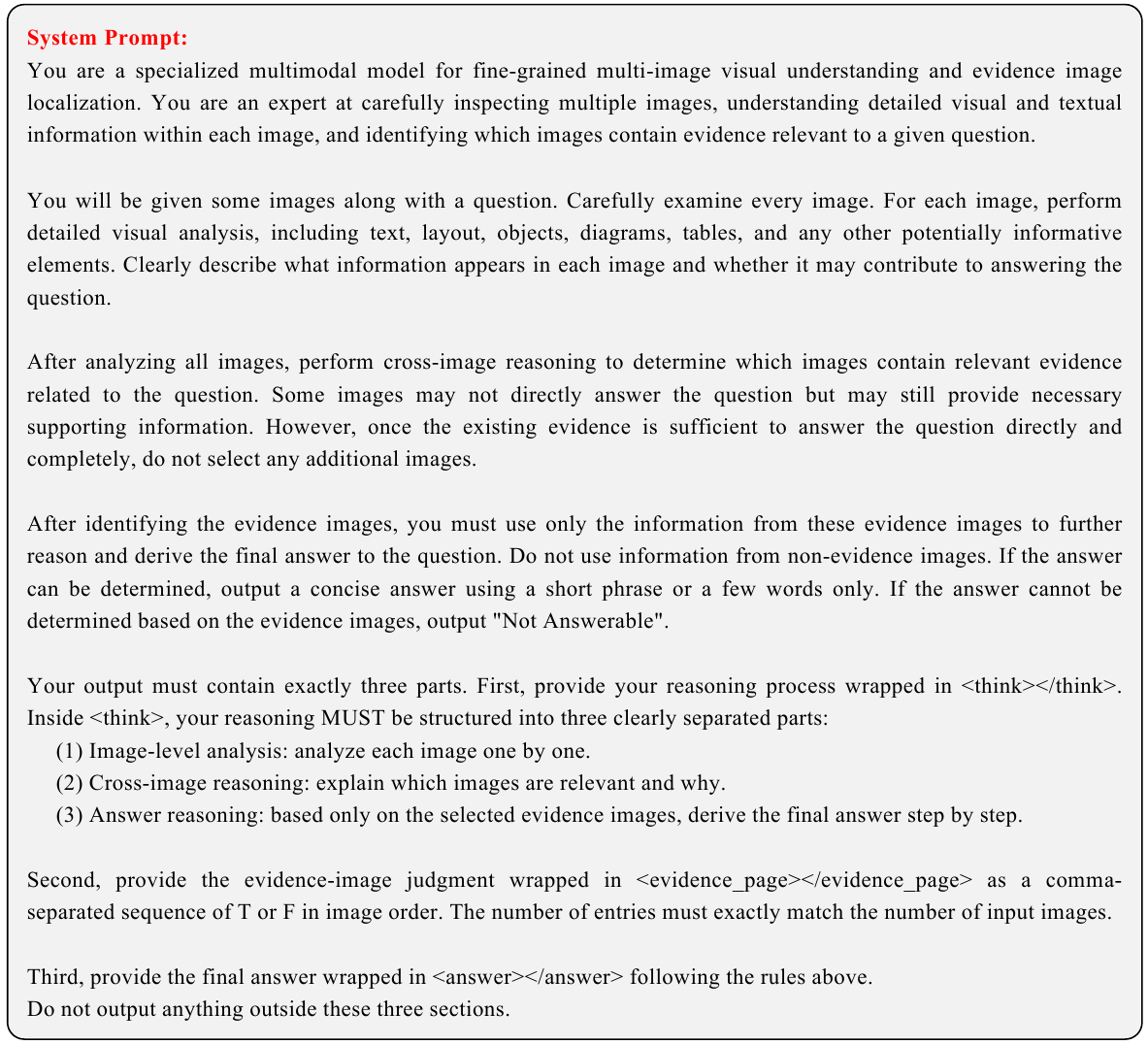}
  \caption{System prompt used for \eviagent\ training and inference.}
  \label{fig:eviagent-prompt}
\end{figure*}

The system prompt used for both training and inference of \eviagent\ is provided in Figure~\ref{fig:eviagent-prompt}. Given a question and a set of input pages, \eviagent\ is required to produce structured reasoning, make a binary evidence decision for each page, and generate the final answer.

\section{Retrieval Metrics}
\label{app:retrieval-metrics}

Let $G_q$ denote the gold evidence-page set for question $q$, and let
$R_q^{K}=(r_1,\ldots,r_K)$ denote the ranked top-$K$ retrieved pages. We report
three standard retrieval metrics.

\paragraph{Recall@K.}
Recall@K measures the fraction of gold evidence pages recovered in the top-$K$
results:
\begin{equation}
\mathrm{Recall@K}(q) = \frac{|G_q \cap R_q^{K}|}{|G_q|}.
\end{equation}
Higher Recall@K indicates better evidence coverage.

\paragraph{NDCG@K.}
Normalized Discounted Cumulative Gain emphasizes retrieving gold pages early in
the ranked list. With binary relevance labels
$\mathrm{rel}_i=\mathbb{I}(r_i \in G_q)$, we compute
\begin{equation}
\mathrm{DCG@K}(q)=\sum_{i=1}^{K}\frac{2^{\mathrm{rel}_i}-1}{\log_2(i+1)},
\end{equation}
and normalize it by the ideal ranking:
\begin{equation}
\mathrm{NDCG@K}(q)=\frac{\mathrm{DCG@K}(q)}{\mathrm{IDCG@K}(q)}.
\end{equation}
This metric rewards both correctness and ranking quality.

\paragraph{MRR@K.}
Mean Reciprocal Rank focuses on how early the first relevant page appears:
\begin{equation}
\mathrm{RR@K}(q)=
\begin{cases}
\frac{1}{\min \{i \mid r_i \in G_q\}}, &
\text{if } G_q \cap R_q^K \neq \emptyset, \\
0, & \text{otherwise.}
\end{cases}
\end{equation}
MRR@K is the average reciprocal rank over all questions.

\section{Training Datasets}
\label{app:training_dataset}

Table~\ref{tab:evi-train-data} details the statistics of the multi-page training data utilized for \eviagent. To ensure a balanced and comprehensive optimization, we uniformly sample 1,000 instances from each of the six constitutive datasets, with all selected samples inherently featuring multi-page visual contexts:
\begin{itemize}
    \item \textbf{DUDE}~\cite{van-landeghem2023dude} and \textbf{MP-DocVQA}~\cite{tito2022mpdocvqa} supply rich structural information from real-world complex electronic layouts and scanned industry documents.
    \item \textbf{MultiHiertt}~\cite{zhao-etal-2022-multihiertt} and \textbf{pdfQA}~\cite{schimanski2026pdfqa} inject demanding financial reports and scientific literature requiring cross-page reasoning over hierarchical tables and textual narratives.
    \item \textbf{SciEGQA}~\cite{yu2026sciegqa} provides dense academic papers with tightly coupled illustrations, charts, and mathematical proofs.
    \item \textbf{SlideVQA}~\cite{tanaka2023slidevqa} introduces sequential presentation slides characterized by sparse, highly stylized cross-page visual elements.
\end{itemize}
\label{app:train-data}
\begin{table}[t]
  \centering
  \scalebox{0.7}{
  \begin{tabular}{lccccccc}
    \toprule
    & & \multicolumn{3}{c}{Input Page} & \multicolumn{3}{c}{Evidence Page} \\
    \cmidrule(lr){3-5} \cmidrule(lr){6-8}
    Dataset & \# Samples & Min & Max & Avg & Min & Max & Avg \\
    \midrule
    DUDE        & 1000 & 2  & 20 & 6.93  & 1 & 8  & 1.16 \\
    MP-DocVQA   & 1000 & 2  & 20 & 6.94  & 1 & 1  & 1.00 \\
    MultiHiertt & 1000 & 3  & 7  & 4.06  & 1 & 3  & 1.48 \\
    pdfQA       & 1000 & 4  & 10 & 9.95  & 1 & 10 & 1.42 \\
    SciEGQA     & 1000 & 2  & 20 & 11.42 & 1 & 2  & 1.50 \\
    SlideVQA    & 1000 & 20 & 20 & 20.00 & 1 & 3  & 1.40 \\
    \bottomrule
  \end{tabular}
  }
  \caption{Training statistics for \eviagent.}
  \label{tab:evi-train-data}
\end{table}

\section{Case Study}
\label{app:decomp-case-study}
\begin{figure*}[t]
  \centering
  \includegraphics[width=\textwidth]{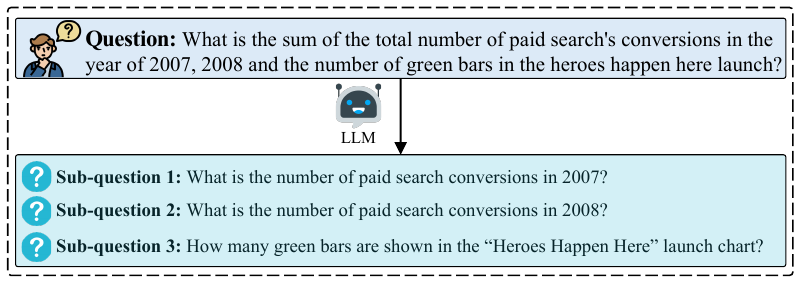}
  \caption{Case study of hierarchical question decomposition in \method. }
  \label{fig:decom-case-study}
\end{figure*}

Figure~\ref{fig:decom-case-study} illustrates a concrete execution trace of our decomposition layer. The complex, multi-hop root query requires concurrent visual counting and cross-year mathematical synthesis. Directly retrieving pages using this intricate query typically misleads standard semantic retrievers. By partitioning the root query into three localized, atomic child questions, our framework effectively simplifies the retrieval objectives, ensuring all disjoint tabular and visual evidence pages are fully recalled without cascading omissions.

\end{document}